\documentclass[conference]{IEEEtran}
\usepackage{cite}
\usepackage{amsmath,amssymb,amsfonts}
\usepackage{algorithmic}
\usepackage{graphicx}
\usepackage{textcomp}
\usepackage{xcolor}
\usepackage[T1]{fontenc}
\usepackage[utf8]{inputenc}
\usepackage{bm}
\usepackage{booktabs}
\usepackage{tabularx}
\usepackage{enumitem}
\usepackage[hidelinks]{hyperref}
\title{Improving X-Ray Baggage Instance Segmentation with Shared Point-Based Mask Refinement}
\author{
\IEEEauthorblockN{Abdulaziz Alqahtani\textsuperscript{1,*}, and Qasim Umer\textsuperscript{1,2}}
\IEEEauthorblockA{\textsuperscript{1}Department of Information and Computer Science, King Fahd University of Petroleum \& Minerals, Saudi Arabia\\
\textsuperscript{2}Interdisciplinary Research Center for Intelligent Secure Systems, KFUPM, Saudi Arabia\\
\textsuperscript{*}Corresponding author: Abdulaziz Alqahtani (\href{mailto:s202283240@kfupm.edu.sa}{s202283240@kfupm.edu.sa})\\
Qasim Umer: \href{mailto:Qasim.Umer@kfupm.edu.sa}{Qasim.Umer@kfupm.edu.sa}}
}
\begin{document}
\maketitle\begin{abstract}

Overlapping projections and complex object contours make prohibited items difficult to delineate in X-ray baggage images. This study investigates shared point-based mask refinement to improve instance segmentation within Cascade Mask R-CNN. The framework combines a ResNet-101 backbone with an attention-based feature pyramid and progressively updates uncertain mask locations, increasing mask resolution from 28 \ensuremath{\times} 28 to 112 \ensuremath{\times} 112. A single refinement module is shared across cascade stages and refinement levels, with transformer and multilayer perceptron (MLP) variants used to examine the role of interactions among selected points. Ten configurations were evaluated on the official PIXray split across three training seeds using the original target-instance annotations for evaluation. The transformer and MLP refiners achieved mean mask average precision (AP) of 71.79 and 72.47, respectively, compared with 70.50 for the unrefined detector, yielding gains of 1.29 and 1.97 AP points. The MLP achieved the highest mean mask AP among the evaluated configurations. Both variants also improved boundary agreement on commonly detected objects, supporting improved mask delineation, while detection diagnostics did not establish a corresponding positive gain. These findings demonstrate that shared point-based refinement can improve segmentation quality in the tested cascade architecture using existing target-mask annotations, without additional occluder supervision. Evaluation on independent X-ray datasets is needed to determine how well these improvements generalize.

\end{abstract}
\begin{IEEEkeywords}X-ray baggage inspection; prohibited items; instance segmentation; point-based mask refinement; uncertainty-guided sampling; PIXray.\end{IEEEkeywords}

\section{Introduction}\label{introduction}

Security screening, including X-ray baggage inspection at airports and public venues, plays an important role in detecting prohibited items and maintaining public safety. Automated detection and segmentation can support screening by helping identify threats in complex baggage images~\cite{ref1}. However, X-ray imagery presents challenges that differ from conventional photographs. Objects overlap in projection, their appearances vary with material properties, and clutter can obscure the features needed to distinguish prohibited items from surrounding objects~\cite{ref2,ref3}.

Object detection identifies an item through its class and bounding box, whereas instance segmentation additionally estimates its pixel-level extent. Mask R-CNN~\cite{ref4} provides a foundation for joint detection and segmentation, and Cascade R-CNN~\cite{ref5} progressively refines object localization. X-ray-specific approaches further address overlap through dense de-overlap attention~\cite{ref3} and occlusion-aware bilayer modeling~\cite{ref6}. A correct detection box, however, does not guarantee an accurate mask. In the detector studied here, each instance is initially represented by a coarse 28 \ensuremath{\times} 28 mask grid, so where projected objects overlap or have thin parts, the mask can omit parts of the object or include neighboring material.

Uncertainty-guided refinement allocates additional prediction capacity to selected locations that may require more precise delineation, including locations near boundaries; uncertainty is a selection heuristic and does not identify every mask error. PointRend~\cite{ref7} established this form of point-based refinement, re-predicting labels at selected points from image features and coarse-mask information instead of recomputing the whole mask at higher resolution. RefineMask~\cite{ref8} progressively incorporates fine-grained features, and Mask Transfiner~\cite{ref9} uses transformer-based processing to refine error-prone regions. Building on this work, our contribution is the shared, progressive integration of point refinement across the stages of a Cascade Mask R-CNN with ResNet-101 and AttFPN~\cite{ref2}, referred to here as the Parent, together with a controlled comparison of two point processors.

We test whether this module improves instance masks under unchanged target-mask supervision, whether interaction among selected points improves performance relative to point-wise processing, and whether any gains reflect mask delineation rather than increased detection performance. PIXray~\cite{ref3} provides a common benchmark for these questions because its official split includes both bounding boxes and instance masks for the same objects, and its overlapping items and varied object geometry are relevant to refinement. It is one dataset, however, and its all-positive test set cannot assess rejection of clean baggage or establish generalization across scanners or sites.

The study addresses three research questions:

\begin{itemize}
\item
  \textbf{RQ1:} Does shared point refinement improve mask AP over the Parent?
\item
  \textbf{RQ2:} How does attention among selected points affect mask AP relative to point-wise processing, and are the two variants equivalent within the study's tolerance?
\item
  \textbf{RQ3:} Does point refinement improve mask delineation, as assessed by stricter mask-overlap criteria and boundary agreement?
\end{itemize}

The corresponding hypotheses are:

\begin{itemize}
\item
  \textbf{H1:} Each refiner achieves a positive mask-AP gain over the Parent.
\item
  \textbf{H2:} The mean mask-AP difference between the MLP and transformer refiners lies within the study's equivalence bounds of \ensuremath{-}1 to +1 AP point.
\item
  \textbf{H3:} Relative to the Parent, point refinement produces a larger gain in mask AP\textsubscript{75} than in mask AP\textsubscript{50}.
\item
  \textbf{H4:} Relative to the Parent, point refinement improves boundary IoU on objects matched by both compared experiments in every evaluated seed.
\end{itemize}

H3 and H4 are evaluated separately for the transformer and MLP refiners. Box AP and detection coverage were examined after scoring as diagnostic analyses to assess whether the mask-quality gains were accompanied by changes in detection performance.

\section{Related work}\label{related-work}

X-ray benchmarks differ in what they supervise, which matters for interpreting segmentation gains. PIDray~\cite{ref2} supports classification, detection and instance segmentation, whereas PIXray~\cite{ref3} provides instance masks for 15 categories. LDXray~\cite{ref10} and the balanced dataset of Zhang et al.~\cite{ref11} address dual-view and class-balanced detection. Ren et al.~\cite{ref6} added occluder annotations to create PIDray-A and PIXray-A, enabling explicit supervision of occluding regions. A gain obtained with this additional supervision answers a different question from a gain obtained with the original target masks alone, which is the setting of our refiners. For overlapping items, Zhang et al.~\cite{ref12} proposed weakly supervised semantic segmentation from category labels, and Nasim et al.~\cite{ref13} studied incremental instance segmentation with knowledge distillation. Detection methods such as AO-DETR~\cite{ref14}, MMCL~\cite{ref15} and surface-aware detection~\cite{ref16} reduce feature interference among overlapping items, but box-level results do not establish mask quality. CLIP adaptation~\cite{ref17} addresses open-vocabulary X-ray detection rather than masks.

Foundation-model adaptation and coarse-mask refinement offer complementary approaches to segmentation. RSPrompter~\cite{ref18} learns prompts for SAM-based instance segmentation, and SAM 2~\cite{ref19} provides related promptable image and video segmentation as contextual literature rather than an evaluated baseline. In the X-ray setting, Ren et al.~\cite{ref6} combine SAM representations with separate occluder and target decoders under occluder supervision. Coarse-mask refinement instead retains the detector architecture and corrects its mask: PointRend~\cite{ref7} re-predicts uncertain points with a point-wise head, RefineMask~\cite{ref8} progressively adds fine-grained features, Mask Transfiner~\cite{ref9} applies attention among error-prone points, and SAMRefiner~\cite{ref20} derives prompts to refine existing masks outside the X-ray domain.

Prior work has examined processor choice: Mask Transfiner~\cite{ref9} compared transformer and MLP refinement within its quadtree framework and reported an advantage for transformer processing. Our question concerns whether that advantage transfers to shared progressive refinement in a cascade detector for X-ray images. We compare closely matched processors within an otherwise shared refiner (Section~\ref{uncertainty-guided-mask-refinement}). The reference experiments are adaptations to a common protocol, and no established refinement method is reproduced directly; the findings therefore concern the integrated refiner and processor comparison rather than superiority over existing refinement systems.

\section{Methodology}\label{methodology}

\subsection{\texorpdfstring{Dataset }{Dataset }}\label{dataset}

We conducted our experiments on PIXray~\cite{ref3}, which contains pseudo-colored X-ray baggage images annotated for 15 prohibited-item categories. We used the official training and test split without modification: 3,560 training images containing 10,709 annotated instances and 1,486 test images containing 4,508 instances, giving 5,046 images and 15,217 annotated instances in the annotation files used in this campaign. Every image contains at least one annotated prohibited item. The released annotations provide, for each object, a category label, a bounding box and a polygon-based instance mask, which serves as its target mask. We retained these annotations unchanged; Figure 1 shows examples of each category, and Table 5 gives per-class test counts.

PIXray-A~\cite{ref6} adds occluder annotations to PIXray, marking the region of the overlapping object. Only the occlusion-aware bilayer SAM and bilayer FCN baselines (Section~\ref{compared-baselines}) additionally use these PIXray-A occluder masks, and only during training; all other models, including both proposed refinement variants, are trained with the original target-instance masks. All models are evaluated against the original PIXray target-instance annotations of the official test split.

\begin{figure*}[tb]
\centering
\includegraphics[width=.90\linewidth,height=.66\textheight,keepaspectratio]{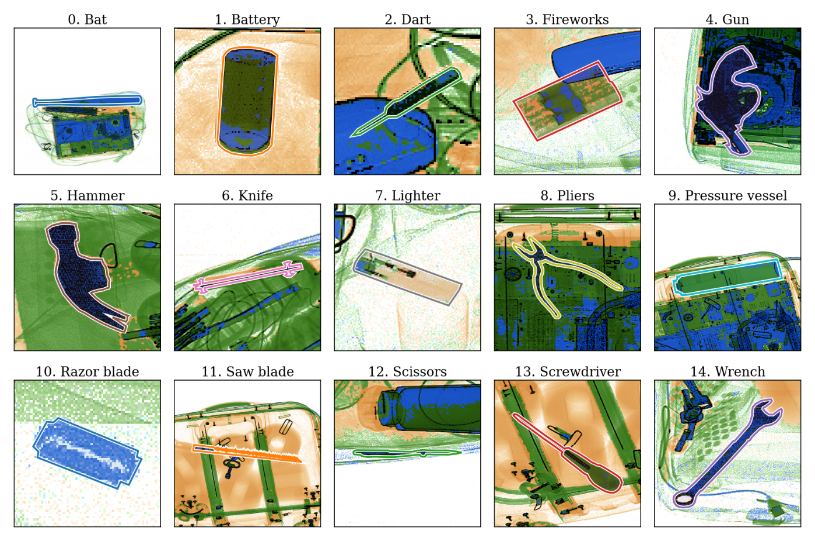}
\caption{Training-set examples of the 15 PIXray categories with ground-truth mask outlines.}
\end{figure*}

\subsection{Data preprocessing and augmentation}\label{data-preprocessing-and-augmentation}

All models use the same augmentation settings, with random transformations sampled independently for each training image. Images are flipped horizontally with probability 0.5, resized with the aspect ratio preserved at a nominal 512 \ensuremath{\times} 512 scale multiplied by a ratio sampled from {[}0.5, 1.5{]}, and randomly cropped to at most 512 \ensuremath{\times} 512; crops without instances are permitted. Boxes are recomputed from the transformed masks, empty instances are removed, and occluder masks, where used, receive the same geometric transformations. Images are converted from BGR to RGB and normalized with channel means (123.675, 116.28, 103.53) and standard deviations (58.395, 57.12, 57.375) in input pixel units. Training batches are padded to their largest image dimensions, rounded up to multiples of 32, and the SAM-based baselines additionally pad to a fixed 512 \ensuremath{\times} 512.

At inference, images are resized with the aspect ratio preserved to fit within 512 \ensuremath{\times} 512 and padded to that size. No test-time augmentation is used, and predictions are mapped back to the original image coordinates for evaluation.

\subsection{Base instance-segmentation framework}\label{base-instance-segmentation-framework}

The Parent, Cascade Mask R-CNN with AttFPN, combines Cascade Mask R-CNN~\cite{ref5} with an ImageNet-pretrained ResNet-101 backbone~\cite{ref21} and AttFPN~\cite{ref2} (Figure 2). The backbone stem and first residual stage are frozen, and batch-normalization running statistics remain fixed during training. AttFPN adds channel and spatial attention to a standard feature pyramid~\cite{ref22} and produces five 256-channel levels, P2--P6, with strides 4--64; the finest level, P2, supplies the point refiner.

\begin{figure*}[tb]
\centering
\includegraphics[width=.92\linewidth,height=.66\textheight,keepaspectratio]{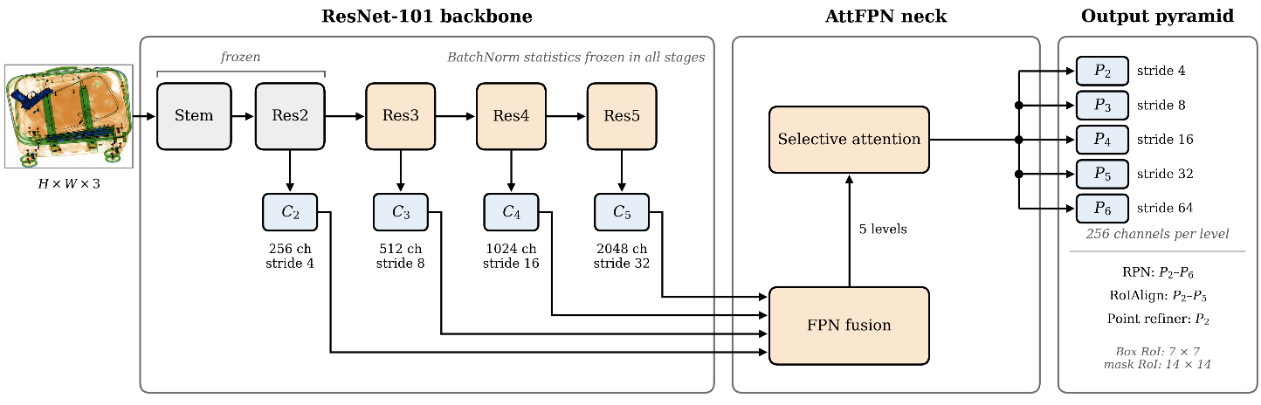}
\caption{ResNet-101 backbone and AttFPN feature hierarchy.}
\end{figure*}

The three cascade stages use assignment IoU thresholds of 0.5, 0.6 and 0.7, and each stage has its own mask head that maps region-of-interest (RoI) features to class-specific coarse-mask logits:

\[F_{RoI}^{(s)} \in \mathbb{R}^{K_{s} \times 256 \times 14 \times 14}\  \longrightarrow \ M^{(s)} \in \mathbb{R}^{K_{s} \times 15 \times 28 \times 28}\]

where \(K_{s}\) is the number of positive training RoIs at stage \(s\) and may differ across stages; it is not the number of final detections at inference, when the heads are applied to the final detection RoIs. The output contains coarse-mask logits for the 15 PIXray categories.

Training combines the region proposal network (RPN) losses with cascade classification, box-regression and coarse-mask losses; the cascade losses use stage weights of 1, 0.5 and 0.25. Coarse masks use binary cross-entropy supervision on the ground-truth class channel of positive RoIs~\cite{ref4}. The Parent also supplies the final detection boxes and classes: at inference, stage classification scores are averaged, and final-stage boxes undergo class-wise non-maximum suppression. For each final detection, the corresponding mask of every stage is refined (Section~\ref{uncertainty-guided-mask-refinement}) and converted to probabilities with a sigmoid; the stage probabilities are averaged, and the predicted-class channel is projected into the detection box in the original image and thresholded at 0.5. The Parent follows the same sequence without the refinement step. Figure 3 summarizes this inference pipeline.

\begin{figure*}[tb]
\centering
\includegraphics[width=.92\linewidth,height=.66\textheight,keepaspectratio]{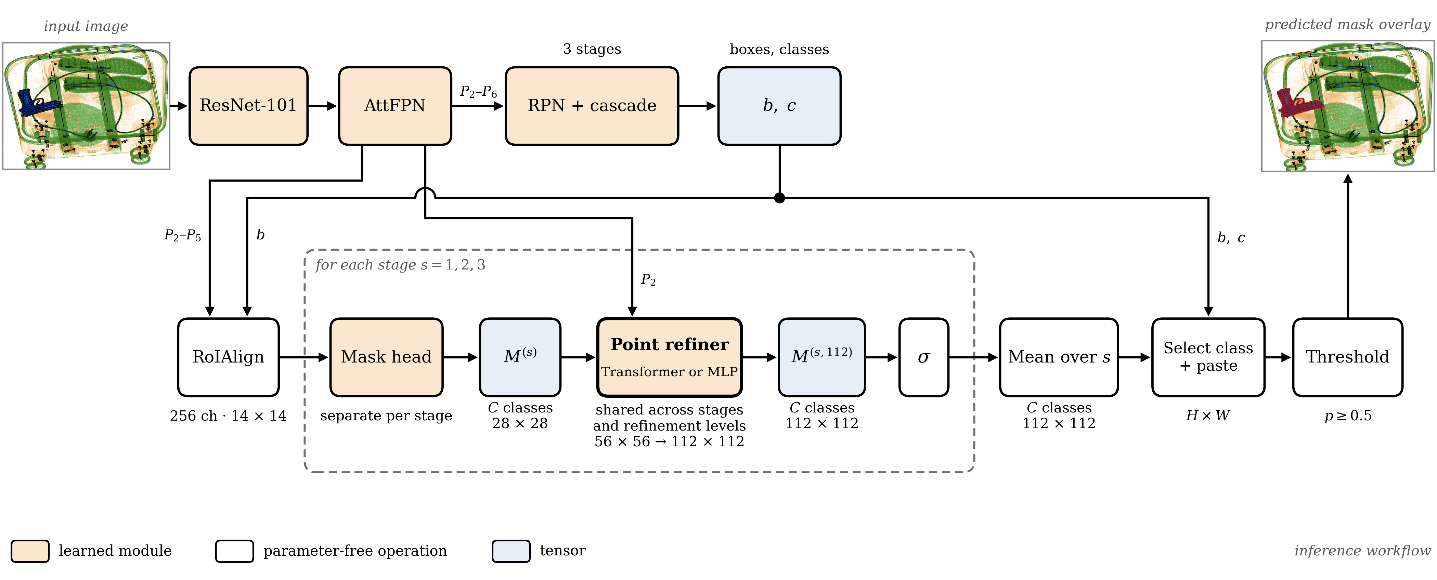}
\caption{Overall inference pipeline with shared point refinement.}
\end{figure*}

\subsection{Uncertainty-guided mask refinement}\label{uncertainty-guided-mask-refinement}

Followed by PointRend~\cite{ref7}, we extend the Parent with a shared point-refinement module that refines uncertain locations in coarse instance masks, progressively refines the 28 \ensuremath{\times} 28 coarse masks of each cascade stage to 56 \ensuremath{\times} 56 and then 112 \ensuremath{\times} 112, using image features, coarse-mask predictions, point coordinates and the refinement level. Each level updates logits at selected uncertain locations, retaining interpolated values elsewhere to produce a dense mask. The refiner is shared across all three cascade stages and both refinement levels; coarse mask heads remain separate.

We investigate two separately trained variants: Transformer point refinement (Figure 4), which allows interactions among selected points within an RoI, and MLP point refinement (Figure 5), which processes points independently. Only the point processor differs; point selection, input embeddings, decoder, refinement levels, supervision, and the Parent architecture are shared. Each variant adds approximately 0.46 million parameters (Table 6) and is trained jointly with the detector.

\subsubsection{Point selection and embedding}\label{point-selection-and-embedding}

At each refinement level, the current mask of each RoI is first upsampled to the level\textquotesingle s resolution. Point selection uses this evolving upsampled mask, which at the second level already contains the first level\textquotesingle s updates. For a cascade stage \(s\), a refinement resolution \(r\) and a RoI-relative location \(u\), the uncertainty is

\begin{equation}
U^{(s,\ r)}(u) = - \left| {\widetilde{M}}_{c}^{(s,\ r)}(u) \right|,\quad r \in \left\{ 56,\ 112 \right\}
\end{equation}

where \(\widetilde{M}\) denotes the upsampled logits before the current level\textquotesingle s update, and the class \(c\) is the assigned ground-truth class during training and the predicted class during inference. Logits closer to zero indicate greater uncertainty; second level therefore incorporates the first-level updates.

Training selects 196 unique points per positive RoI and level: the 147 most uncertain among 588 uniformly sampled candidates, plus 49 uniformly sampled from the remaining grid locations. Inference selects 784 points per RoI and level with the largest adjusted scores \(U^{(s,r)}(u_j)-j\epsilon\), where \(j\) is the zero-based flattened grid index and \(\epsilon\) is the FP32 machine epsilon. This index-dependent offset can reorder near-equal uncertainty values. Sampling is without replacement at grid-cell centers, with normalized coordinates \(x = \frac{(column + 0.5)}{W},\ \ \ y = \frac{(row + 0.5)}{H}\). These RoI-relative coordinates are converted to augmented-image coordinates before features and targets are sampled. This discrete sampling adapts the continuous training-point sampling used by PointRend and permits direct replacement at the selected grid locations.

Embedding then samples the original 28 \ensuremath{\times} 28 coarse logits of the same stage, not the evolving mask. Each selected point becomes a 128-dimensional token formed from four inputs: image features, original coarse-mask logits, RoI-relative coordinates and a refinement-level embedding,

\begin{equation}
z_{i} = LN\left\lbrack \varphi_{f}\left( f_{i} \right) + \varphi_{m}\left( m_{i} \right) + \varphi_{p}\left( u_{i} \right) + e_{r} \right\rbrack \in \mathbb{R}^{128}
\end{equation}

where \(f_{i}\  \in \ \mathbb{R}^{256}\) is the P2 feature sampled bilinearly at point \(i\), \(m_{i}\  \in \ \mathbb{R}^{15}\) are the logits of all classes sampled from the stage\textquotesingle s original coarse mask, \(u_{i}\  \in \ \lbrack 0,\ 1\rbrack^{2}\) are normalised RoI-relative coordinates, \(e_{r}\  \in \ \mathbb{R}^{128}\) is a learned refinement-level embedding and LN is layer normalization. The image branch uses a linear 256 to 128 projection. The mask and coordinate branches use two linear layers separated by GELU, with dimensions 15 \ensuremath{\rightarrow} 64 \ensuremath{\rightarrow} 128 and 2 \ensuremath{\rightarrow} 128 \ensuremath{\rightarrow} 128, respectively. The four resulting 128-dimensional terms are summed. Projection weights are shared across stages and levels, and no cascade-stage embedding is used.

\subsubsection{Refinement variants and mask update}\label{refinement-variants-and-mask-update}

For one RoI the tokens are stacked into \(Z^{(0)}\  = \ \left\lbrack {z_{1}}^{\top};\ \ldots\ ;\ {z_{P}}^{\top} \right\rbrack\  \in \ \mathbb{R}^{P \times 128}\), where \(P\) is the point budget per RoI and level is, and passed through two pre-normalization residual blocks:

\begin{equation}
\begin{aligned}
H^{(b)} & = Z^{(b\  - \ 1)} + {Mix}_{b}\left( {LN}_{b,\ 1}\left( Z^{(b\  - \ 1)} \right) \right), \\
Z^{(b)} & = H^{(b)} + {FFN}_{b}\left( {LN}_{b,\ 2}\left( H^{(b)} \right) \right),\quad b \in \left\{ 1,\ 2 \right\}
\end{aligned}
\end{equation}

Here, b indexes the two sequential processor blocks. The two variants differ only in Mix. In Transformer point refinement, Mix is multi-head self-attention with four heads of 32 channels each, and attention operates only among the points of the same RoI. In MLP point refinement, Mix is a point-wise 128 \ensuremath{\rightarrow} 256 \ensuremath{\rightarrow} 128 MLP with GELU, so points are processed independently. Both variants use the same 128 \ensuremath{\rightarrow} 512 \ensuremath{\rightarrow} 128 feed-forward network with GELU, layer normalization and residual connections, and no dropout. The two sequential blocks have separate parameters, whereas the complete refiner is reused across cascade stages and refinement levels.

A final layer normalization and a linear 128 \ensuremath{\rightarrow} 15 decoder give class logits at each selected point. These logits replace the interpolated values at the selected locations rather than being added as residual corrections. The updated 56 \ensuremath{\times} 56 grid is then upsampled and refined again to produce the 112 \ensuremath{\times} 112 grid. At inference, the refined masks of each stage enter the probability averaging and projection described in Section~\ref{base-instance-segmentation-framework}.

\begin{figure*}[tb]
\centering
\includegraphics[width=.92\linewidth,height=.66\textheight,keepaspectratio]{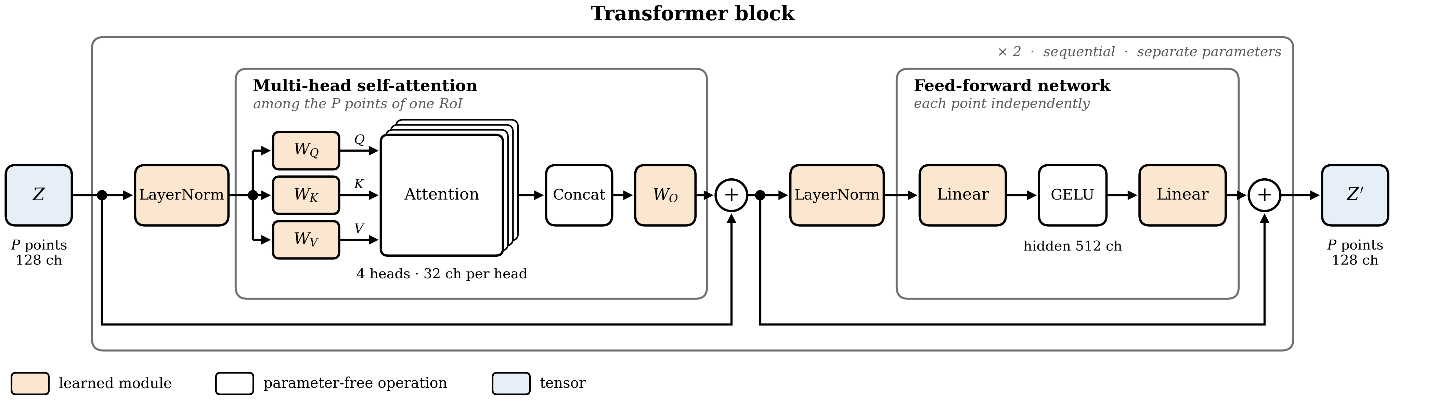}
\caption{Transformer point processor with two sequential blocks and separate parameters.}
\end{figure*}

\begin{figure*}[tb]
\centering
\includegraphics[width=.92\linewidth,height=.66\textheight,keepaspectratio]{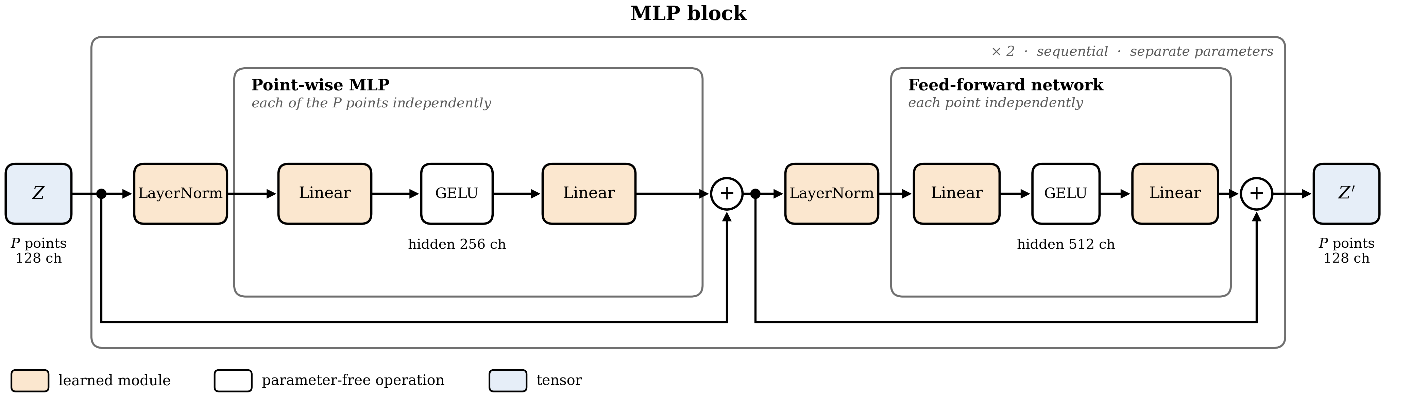}
\caption{MLP point processor with two sequential blocks and separate parameters.}
\end{figure*}

\subsection{Training objective}\label{training-objective}

Point supervision supplements the base-detector losses and applies to positive RoIs at both refinement levels of every cascade stage. Targets are sampled bilinearly from the augmented ground-truth instance masks before their reduction to the coarse 28 \ensuremath{\times} 28 RoI targets, so interpolation can yield soft targets in {[}0, 1{]}, especially near boundaries. For stage \(s\) and level \(r\),

\begin{multline}
L_{point}^{s,r}=\frac{1}{K_sP}\sum_{k=1}^{K_s}\sum_{i\in S_{s,k}^{r}}\\[-0.25ex]
\operatorname{BCEWithLogits}\!\bigl(\ell_{s,k,i}^{r}[c_{s,k}],\,g_{s,k}(p_{s,k,i}^{r})\bigr)
\end{multline}

where \(K_{s}\) is the number of positive RoIs, indexed by \emph{k}, at stage \(s\), \(P\) is the training point budget of 196 points per RoI and level, \(S_{s,\ k}^{r}\) is the selected point set, \(c_{s,\ k}\) is the assigned ground-truth class, \(\ell_{s,\ k,\ i}^{r}\) is the decoded logit vector at point \(i\), from which the bracket selects the assigned class, and \(g_{s,\ k}\) is the assigned augmented ground-truth instance mask evaluated at \(p_{s,\ k,\ i}^{r}\) in augmented-image coordinates. The expression applies when \(K_{s}\  > \ 0\); otherwise, the stage contributes no point loss.

Averaging the two levels gives the complete objective,

\begin{equation}\begin{aligned}
&L = L_{base} + \frac{1}{2}\sum_{s = 1}^{3}{w_{s}\left( L_{point}^{s,\ 56} + L_{point}^{s,\ 112} \right)},\\
&\left( w_{1},\ w_{2},\ w_{3} \right) = (1,\ 0.5,\ 0.25)
\end{aligned}\end{equation}

where \(L_{base}\) comprises the existing RPN loss and the weighted cascade classification, box-regression and coarse-mask losses. The two levels contribute equally, stage weights are applied once, and no additional point-loss multiplier is tuned. The coarse-mask loss is unchanged, and the refiner is trained jointly with the detector.

\subsection{Compared baselines}\label{compared-baselines}

The two refiners were compared with eight reference experiments. These comprise Mask R-CNN~\cite{ref4} and Cascade Mask R-CNN~\cite{ref5} with standard feature pyramids; the AttFPN~\cite{ref2} Parent and its recall-RPN and PIDray~\cite{ref2} R0/R1 adaptations; and three SAM or bilayer experiments: RSPrompter~\cite{ref18} with SAM~\cite{ref23} and LoRA~\cite{ref24}, Occ bilayer SAM~\cite{ref6}, and Bilayer FCN~\cite{ref25}.

Only Occ bilayer SAM and Bilayer FCN use additional PIXray-A occluder masks during training. All experiments are evaluated against the original PIXray target annotations under the shared experimental settings. The reference implementations are adaptations to this setting, and the common training budget was not tuned separately for each architecture. Their rankings therefore describe these implementations. The R0/R1 adaptation includes an image-level gate, but the all-positive test set cannot evaluate clean-image rejection.

\section{Experimental setup}\label{experimental-setup}

\subsection{Implementation and training}\label{implementation-and-training}

All experiments were implemented in MMDetection~\cite{ref26} with the shared settings in Table 1, without architecture-specific tuning of the training budget. Training used FP16 automatic mixed precision with dynamic loss scaling; evaluation used FP32 without autocast. Each run was evaluated with its final-epoch checkpoint, without validation-based model selection. Activation checkpointing was enabled in the SAM image encoder of Occ bilayer SAM to reduce training memory. Shared seed labels do not imply identical initialization or augmentation streams across architectures.

\begin{table*}[t]
\footnotesize\centering\caption{Shared experimental settings}
\begingroup\setlength{\tabcolsep}{2pt}
\begin{tabularx}{\textwidth}{>{\raggedright\arraybackslash}p{.26\linewidth} >{\raggedright\arraybackslash}X}
\toprule
\textbf{Setting}
 & \textbf{Value}
 \\
\midrule

Training batch size & 4 \\
Epochs & 50 \\
Optimiser / learning rate & AdamW, 1 \ensuremath{\times} 10\textsuperscript{-4}; weight decay 0.05; betas (0.9, 0.999) \\
Schedule & linear warm-up (25 iterations); epoch-wise cosine decay, configured minimum 1 \ensuremath{\times} 10\textsuperscript{-7} \\
Precision & Training: FP16 AMP with dynamic loss scaling; evaluation: FP32 without autocast \\
Gradient clipping & L2 norm 1.0 \\
Seeds / runs & 42, 123, 2026; 30 training runs \\
Checkpoint rule & final epoch (50); no validation-based selection \\
Inference thresholds & score \ensuremath{\geq} 0.05; NMS IoU 0.5; \ensuremath{\leq} 100 detections; mask threshold 0.5 \\
Hardware & 1 \ensuremath{\times} NVIDIA RTX 4090, 24 GB \\
\bottomrule\end{tabularx}\endgroup
\end{table*}

\subsection{Statistical analysis}\label{statistical-analysis}

Experiments were evaluated on the 1,486 official PIXray~\cite{ref3} test images using the original target annotations, after mapping predictions retained at scores \ensuremath{\geq} 0.05 to original image coordinates. The primary outcome was official COCOeval~\cite{ref27} mask AP, averaged across IoU thresholds of 0.50--0.95 and 15 categories. Results are means over seeds 42, 123, and 2026, with sample SD for primary mask AP; paired effects are means of within-seed differences. AP confidence intervals use 2,000 paired seed-and-image bootstrap replicates~\cite{ref28}. Primary mask-AP differences are tested using 5,000 two-sided paired image permutations~\cite{ref29}, preserving image pairing and the same swap pattern across seeds. Permutation tests condition on the trained models, whereas bootstrap intervals also reflect variation among the observed seeds. H2 compares the 90\% bootstrap interval with \ensuremath{\pm}1-AP equivalence bounds, a study-specific tolerance that has not been externally validated; a separate approximate two one-sided tests (TOST)~\cite{ref30} p-value is reconstructed from the 95\% interval under a normal approximation.

H3 assesses whether the refiner\textquotesingle s gain over the Parent is larger at AP\textsubscript{75} than at AP\textsubscript{50}. H4 assesses the paired boundary-IoU~\cite{ref31} difference on objects matched by both compared experiments in every seed; this reported endpoint replaces the planned Boundary AP, which was not computed. Boundary IoU uses contour bands of 2\% of the image diagonal. Boundary-IoU and detection-coverage analyses use one-to-one, same-class matching at score \ensuremath{\geq}0.5 and box IoU \ensuremath{\geq}0.5; detection coverage includes all ground-truth objects. Image-cluster bootstrap intervals and sign-flip tests account for within-image dependence after averaging object differences across seeds, without resampling seeds. covers separate families: two H1 tests, four joint H3/H4 tests, and 15 per-class tests per refiner--Parent contrast. The H4 endpoint and joint family were specified after the original plan. Positive effect claims require positive estimates, adjusted p \textless{} 0.05, and unadjusted 95\% CIs above zero. Adding two paired mask-IoU tests (six total) left conclusions unchanged at \ensuremath{\alpha} = 0.05. Post-scoring box-AP and detection-coverage diagnostics remain unadjusted and test neither equivalence nor noninferiority. Additional descriptive metrics appear in Appendix~\ref{a.2-additional-descriptive-metrics}.

\section{Results}\label{results}

\subsection{Overall performance and refiner comparison}\label{overall-performance-and-refiner-comparison}

The MLP refiner achieved the highest mean mask AP among the ten experiments, followed by the transformer refiner (Table 2). Both exceeded the Parent under the same original target-mask supervision. Bilayer FCN was the highest-scoring reference experiment, although it and Occ bilayer SAM used additional occluder annotations; reference-experiment rankings are therefore descriptive.

\begin{table*}[t]
\footnotesize\centering\caption{Mean PIXray test performance across ten experiments.}
\begingroup\setlength{\tabcolsep}{2pt}
\begin{tabular*}{\textwidth}{@{\extracolsep{\fill}}l*{6}{c}@{}}
\toprule
\textbf{Experiment}
 & \textbf{Mask AP, mean \ensuremath{\pm} SD}
 & \textbf{95\% CI for mask AP}
 & \textbf{Mask AP\textsubscript{50}}
 & \textbf{Mask AP\textsubscript{75}}
 & \textbf{Mask AR\textsubscript{100}}
 & \textbf{Box AP}
 \\
\midrule

Mask R-CNN & 68.15 \ensuremath{\pm} 0.05 & {[}67.6, 69.1{]} & 93.6 & 78.0 & 71.9 & 77.1 \\
Cascade Mask R-CNN & 70.20 \ensuremath{\pm} 0.04 & {[}69.6, 71.0{]} & 93.9 & 80.7 & 73.2 & 81.1 \\
Cascade + AttFPN (Parent) & 70.50 \ensuremath{\pm} 0.06 & {[}69.9, 71.3{]} & 94.0 & 81.1 & 73.4 & 81.5 \\
Parent + recall RPN & 70.24 \ensuremath{\pm} 0.22 & {[}69.7, 71.1{]} & 94.2 & 80.6 & 73.3 & 81.5 \\
RSPrompter (SAM + LoRA) & 62.45 \ensuremath{\pm} 0.11 & {[}61.9, 63.3{]} & 92.9 & 70.9 & 65.7 & 76.2 \\
Occ bilayer SAM\ensuremath{\dagger} & 62.66 \ensuremath{\pm} 0.10 & {[}62.1, 63.5{]} & 93.0 & 70.8 & 65.9 & 76.1 \\
Bilayer FCN\ensuremath{\dagger} & 70.65 \ensuremath{\pm} 0.08 & {[}70.1, 71.5{]} & 94.3 & 81.4 & 73.7 & \textbf{81.6} \\
Parent + PIDray R0/R1 & 70.36 \ensuremath{\pm} 0.12 & {[}69.8, 71.2{]} & 94.1 & 80.9 & 73.3 & 81.5 \\
Transformer refiner (ours) & 71.79 \ensuremath{\pm} 0.19 & {[}71.1, 72.6{]} & 94.1 & 81.2 & 74.7 & 81.2 \\
MLP refiner (ours) & \textbf{72.47 \ensuremath{\pm} 0.08} & {[}71.9, 73.3{]} & \textbf{94.3} & \textbf{82.4} & \textbf{75.3} & 81.3 \\
\bottomrule\end{tabular*}\endgroup
\par\smallskip{\raggedright Three-seed means (0--100); sample SD across seeds; seed-and-image bootstrap CIs. Bold: highest unrounded column means. \ensuremath{\dagger} Extra occluder supervision.\par}
\end{table*}

Both refiners exceeded the Parent in every seed, supporting H1 with gains of 1.29 AP for Transformer and 1.97 for MLP (both Holm-adjusted p \textless{} 0.001; Table 3; Figure 6). MLP exceeded Transformer by 0.69 AP, but the 90\% CI {[}0.49, 0.88{]} satisfied the study's \ensuremath{\pm}1-AP equivalence criterion. Additional results by object size, mask-matching threshold, and object-level overlap appear in Appendix~\ref{a.2-additional-descriptive-metrics}.

\begin{figure}[tb]
\centering
\includegraphics[width=.94\linewidth,height=.66\textheight,keepaspectratio]{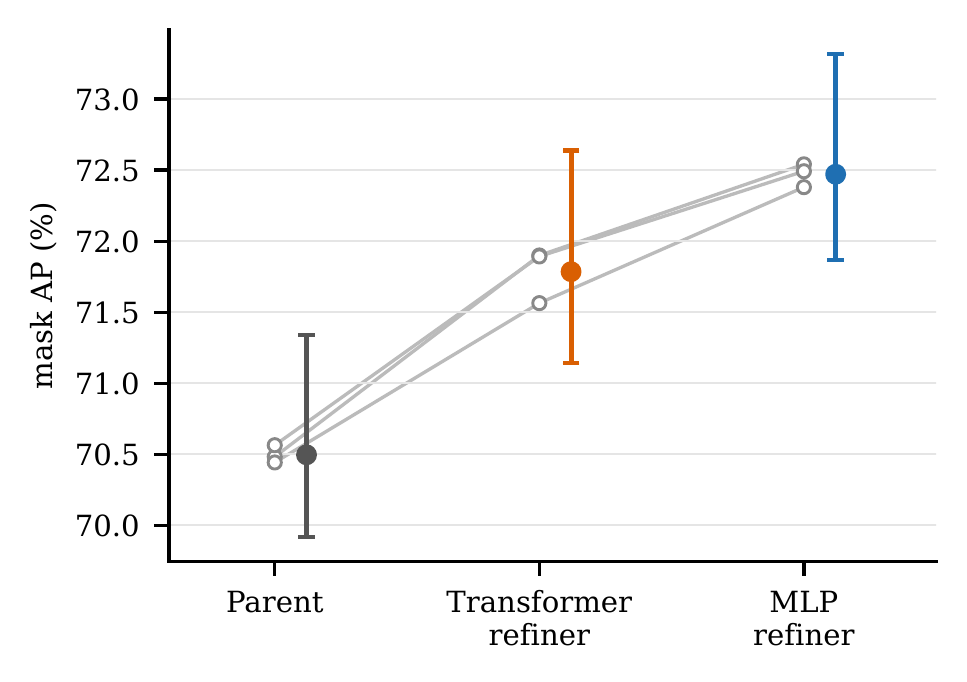}
\caption{Mask AP means and 95\% seed-and-image bootstrap CIs; lines connect same-seed results.}
\end{figure}

\begin{table*}[t]
\footnotesize\centering\caption{Paired mask-AP differences for H1 and H2}
\begingroup\setlength{\tabcolsep}{2pt}
\begin{tabularx}{\textwidth}{>{\raggedright\arraybackslash}p{0.19\linewidth} >{\centering\arraybackslash}p{0.18\linewidth} >{\centering\arraybackslash}p{0.2132\linewidth} >{\centering\arraybackslash}p{0.19\linewidth} >{\centering\arraybackslash}p{0.0969\linewidth} >{\centering\arraybackslash}p{0.0678\linewidth}}
\toprule
\textbf{Hypothesis}
 & \textbf{Paired comparison}
 & \textbf{\ensuremath{\Delta} by seed: 42 / 123 / 2026}
 & \textbf{Mean paired \ensuremath{\Delta} {[}95\% CI{]}}
 & \textbf{Permutation p}
 & \textbf{Holm p}
 \\
\midrule

H1 & Transformer \ensuremath{-} Parent & +1.42 / +1.12 / +1.33 & +1.29 {[}+0.99, +1.54{]} & \textless0.001 & \textless0.001 \\
H1 & MLP \ensuremath{-} Parent & +2.06 / +1.94 / +1.93 & +1.97 {[}+1.75, +2.20{]} & \textless0.001 & \textless0.001 \\
H2: observed difference & MLP \ensuremath{-} transformer & +0.64 / +0.82 / +0.60 & +0.69 {[}+0.45, +0.92{]} & \textless0.001 & --- \\
\bottomrule\end{tabularx}\endgroup
\par\smallskip{\raggedright AP points. CI: paired seed-and-image bootstrap; p: paired image permutation; Holm over the two H1 tests. H2 equivalence: 90\% CI {[}+0.49, +0.88{]} within \ensuremath{\pm}1 AP (approximate TOST p = 0.005). --- not adjusted.\par}
\end{table*}

\subsection{Mask quality and detection diagnostics}\label{mask-quality-and-detection-diagnostics}

The AP\textsubscript{75}-minus-AP\textsubscript{50} gain contrast supported H3 for the MLP refiner (+1.07 AP points; Holm-adjusted p = 0.008) but was inconclusive for the Transformer (+0.11 AP; p = 0.798, Holm-adjusted). Both supported H4 through increased boundary IoU (both Holm-adjusted p \textless{} 0.001; Table 4), but their separate common-match samples prevent direct processor comparison on identical objects. Figure 7's larger gains at stricter mask-IoU thresholds are descriptive and do not replace the H3 test.

All four diagnostic point estimates were slightly negative, and their 95\% confidence intervals included zero. These analyses did not establish either a positive detection gain or unchanged detection performance. For transformer box AP, the percentile-bootstrap interval includes zero before rounding, despite the separate normal-approximation p-value of 0.045; the apparent discrepancy reflects the different procedures and is not treated as a confirmed reduction.

\begin{figure}[tb]
\centering
\includegraphics[width=.94\linewidth,height=.66\textheight,keepaspectratio]{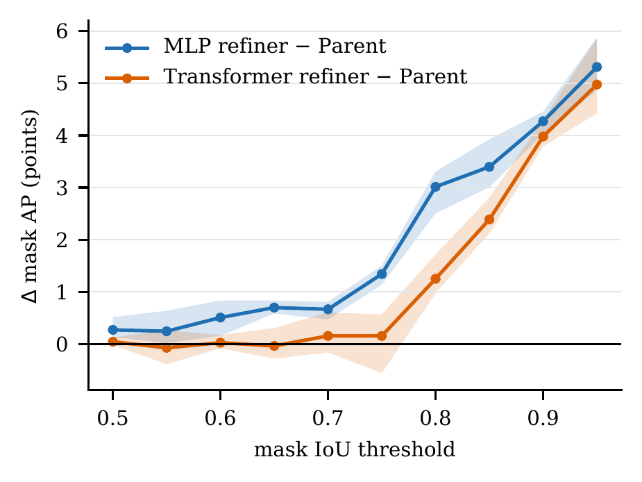}
\caption{Mask-AP gain over the Parent by IoU threshold; shading indicates seed range.}
\end{figure}

\begin{table*}[t]
\footnotesize\centering\caption{Paired changes in mask quality and detection relative to the Parent}
\begingroup\setlength{\tabcolsep}{2pt}
\begin{tabularx}{\textwidth}{>{\raggedright\arraybackslash}p{0.0969\linewidth} >{\centering\arraybackslash}p{0.1551\linewidth} >{\centering\arraybackslash}p{0.1357\linewidth} >{\centering\arraybackslash}p{0.2229\linewidth} >{\centering\arraybackslash}p{0.2035\linewidth} >{\centering\arraybackslash}p{0.0678\linewidth} >{\centering\arraybackslash}p{0.0581\linewidth}}
\toprule
\textbf{Analysis}
 & \textbf{Outcome}
 & \textbf{Refiner}
 & \textbf{Mean \ensuremath{\Delta} {[}95\% CI{]}}
 & \textbf{Test}
 & \textbf{p}
 & \textbf{Holm p}
 \\
\midrule

H3 & \ensuremath{\Delta}AP75 \ensuremath{-} \ensuremath{\Delta}AP50 & Transformer & +0.11 {[}\ensuremath{-}0.81, +0.88{]} & Normal approximation* & 0.798 & 0.798 \\
H3 & \ensuremath{\Delta}AP75 \ensuremath{-} \ensuremath{\Delta}AP50 & MLP & +1.07 {[}+0.30, +1.76{]} & Normal approximation* & 0.004 & 0.008 \\
H4 & Boundary IoU & Transformer & +0.71 {[}+0.64, +0.79{]} & Image-cluster sign-flip & \textless0.001 & \textless0.001 \\
H4 & Boundary IoU & MLP & +1.14 {[}+1.07, +1.21{]} & Image-cluster sign-flip & \textless0.001 & \textless0.001 \\
Diagnostic & Box AP & Transformer & \ensuremath{-}0.27 {[}\ensuremath{-}0.52, +0.00{]} & Normal approximation* & 0.045 & --- \\
Diagnostic & Box AP & MLP & \ensuremath{-}0.18 {[}\ensuremath{-}0.50, +0.15{]} & Normal approximation* & 0.284 & --- \\
Diagnostic & Detection coverage & Transformer & \ensuremath{-}0.22 {[}\ensuremath{-}0.46, +0.01{]} & Image-cluster sign-flip & 0.069 & --- \\
Diagnostic & Detection coverage & MLP & \ensuremath{-}0.17 {[}\ensuremath{-}0.39, +0.05{]} & Image-cluster sign-flip & 0.178 & --- \\
\bottomrule\end{tabularx}\endgroup
\par\smallskip{\raggedright \ensuremath{\Delta}: refiner \ensuremath{-} Parent (AP points; percentage points otherwise). *Normal approximation; Holm: joint H3--H4.\\
AP: 1,486 images \ensuremath{\times} 3 seeds; diagnostics unadjusted; sign-flip tests clustered by image.\par}
\end{table*}

\subsection{Per-class and qualitative results}\label{per-class-and-qualitative-results}

Table 5 reports absolute class AP across all ten experiments. In separate paired comparisons with the Parent, MLP had positive gains in 14 of 15 classes and eight classes met the Holm-adjusted criterion; transformers had positive gains in 11 classes and seven met that criterion. These tests use approximate p-values, with corrections across 15 classes per contrast (for more details, see Appendix Table A1). Dart had the lowest Parent AP and did not improve with either refiner.

\begin{table*}[t]
\footnotesize\centering\caption{Three-seed mean per-class mask AP across ten experiments}
\begingroup\setlength{\tabcolsep}{2pt}
\begin{tabular*}{\textwidth}{@{\extracolsep{\fill}}lr*{10}{c}@{}}
\toprule
\textbf{Class}
 & \textbf{n}
 & \shortstack{\textbf{Mask}\\\textbf{R-CNN}}
 & \textbf{Cascade}
 & \textbf{Parent}
 & \shortstack{\textbf{Recall}\\\textbf{RPN}}
 & \textbf{RSPrompter}
 & \shortstack{\textbf{Occ}\\\textbf{SAM}\ensuremath{\dagger}}
 & \shortstack{\textbf{Bilayer}\\\textbf{FCN}\ensuremath{\dagger}}
 & \textbf{R0/R1}
 & \textbf{Transformer}
 & \textbf{MLP}
 \\
\midrule

Dart & 323 & 37.0 & 37.4 & 38.0 & 38.1 & 20.0 & 20.2 & \textbf{38.8} & 37.8 & 37.7 & 37.7 \\
Saw blade & 287 & 43.2 & 47.8 & 48.2 & 47.8 & 38.7 & 39.0 & 48.5 & 48.5 & 47.8 & \textbf{52.5} \\
Knife & 330 & 53.1 & 56.3 & 56.5 & 56.2 & 45.4 & 45.7 & 56.7 & 56.3 & 57.0 & \textbf{57.9} \\
Screwdriver & 287 & 56.7 & 59.6 & 60.4 & 59.8 & 52.7 & 53.1 & 60.6 & 60.0 & \textbf{63.7} & 63.1 \\
Pliers & 264 & 59.0 & 61.7 & 61.8 & 61.0 & 51.1 & 51.1 & 61.6 & 61.3 & 62.5 & \textbf{66.1} \\
Lighter & 265 & 66.6 & 68.5 & 68.7 & 68.2 & 59.0 & 58.9 & 68.9 & 68.7 & \textbf{69.7} & 69.2 \\
Razor blade & 271 & 69.6 & 69.9 & 70.1 & 70.3 & 61.2 & 61.3 & 70.3 & 69.9 & 70.0 & \textbf{70.3} \\
Wrench & 349 & 67.0 & 70.8 & 71.3 & 71.0 & 66.2 & 66.5 & 71.8 & 71.3 & 73.2 & \textbf{74.2} \\
Scissors & 290 & 73.7 & 75.7 & 75.7 & 75.1 & 62.1 & 63.1 & 75.8 & 75.6 & 77.1 & \textbf{77.3} \\
Fireworks & 282 & 74.8 & 76.5 & 77.2 & 76.7 & 73.9 & 73.8 & 76.6 & 76.8 & \textbf{78.4} & 78.3 \\
Battery & 360 & 79.0 & 80.0 & 80.4 & 80.0 & 74.0 & 73.7 & 80.4 & 80.0 & 80.4 & \textbf{80.4} \\
Gun & 292 & 83.3 & 84.7 & 84.9 & 84.7 & 80.5 & 80.9 & 85.0 & 84.8 & 86.1 & \textbf{86.2} \\
Hammer & 322 & 84.2 & 85.8 & 85.4 & 85.8 & 78.4 & 78.5 & 85.5 & 85.4 & \textbf{86.5} & 86.3 \\
Pressure vessel & 297 & 86.3 & 89.2 & 89.2 & 89.0 & 84.8 & 85.2 & 89.6 & 89.2 & 91.9 & \textbf{92.5} \\
Bat & 289 & 88.7 & 89.2 & 89.7 & 89.8 & 88.8 & 88.9 & 89.6 & 89.6 & 94.7 & \textbf{95.1} \\
\bottomrule\end{tabular*}\endgroup
\par\smallskip{\raggedright Absolute mask AP (\%); n = test objects. Parent = Cascade + AttFPN; Recall RPN and R0/R1 extend the Parent; Occ SAM = Occ bilayer SAM. \ensuremath{\dagger} Extra occluder supervision. Bold: highest unrounded mean (descriptive). For Battery and Razor blade, the MLP has the highest unrounded mean despite ties after rounding.\par}
\end{table*}

Figure 8 shows selected seed-42 successes and failures. Both refiners improved the Bat outline, while the crowded Wrench example shows deterioration, particularly with transformer refinement. The printed values are single-object IoUs from seed 42; these selected examples illustrate the quantitative findings without estimating population-level performance. For more details on the selection rules, see Appendix~\ref{a.1-supporting-numerical-results}.

\begin{figure*}[tb]
\centering
\includegraphics[width=.88\linewidth,height=.66\textheight,keepaspectratio]{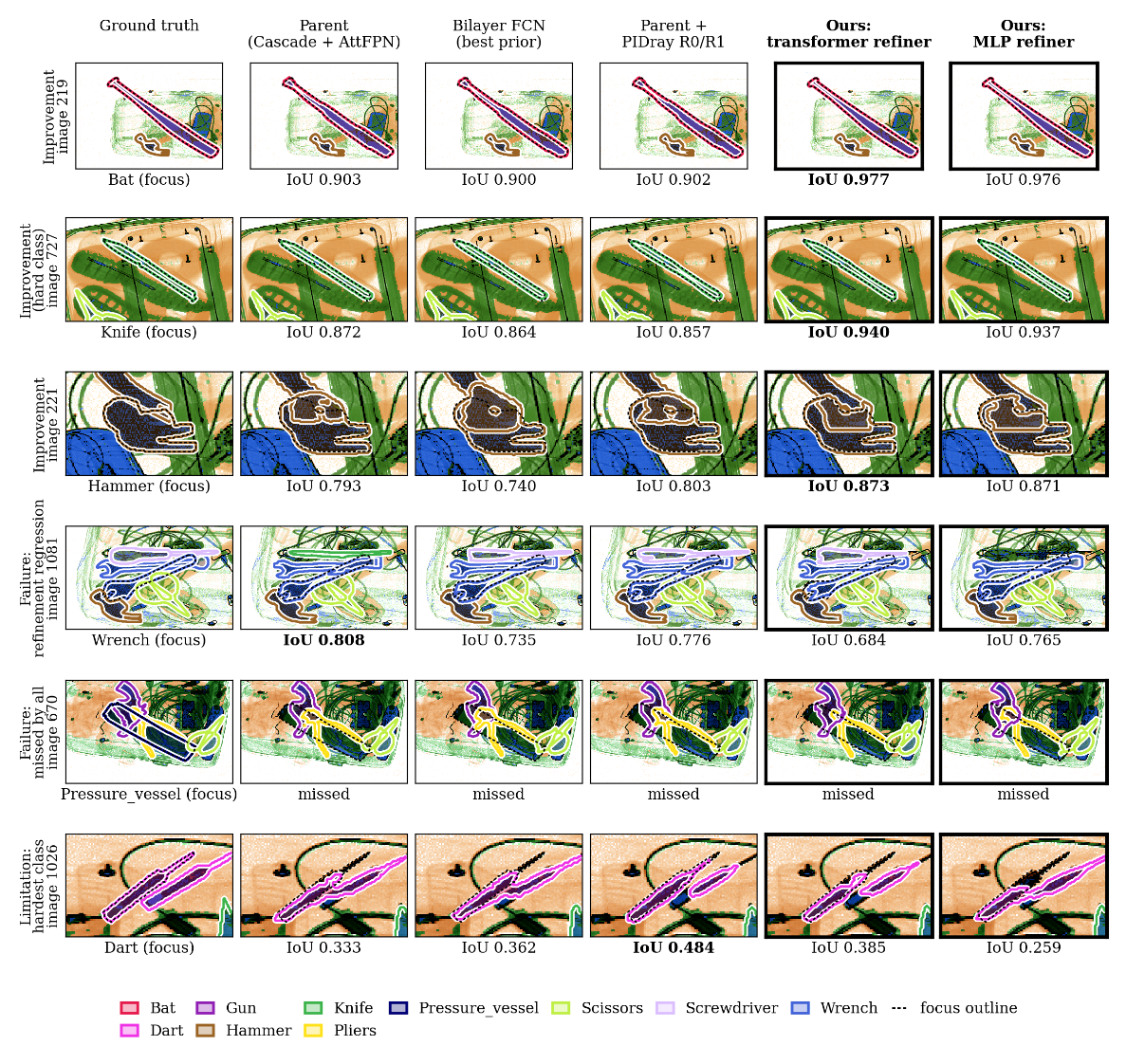}
\caption{Selected refinement successes and failures (seed 42).}
\end{figure*}

``Best prior'' in the image denotes the adapted Bilayer FCN\ensuremath{\dagger} experiment.

\subsection{Computational cost}\label{computational-cost}

The refiners add about 0.46 million parameters to the Parent and required longer mean training, with somewhat lower measured end-to-end throughput (Table 6). The SAM-based experiments have much larger total parameter counts but smaller trainable parameter counts because their encoders are adapted with LoRA. They also have the lowest measured throughput; Occ bilayer SAM\textquotesingle s training also used activation checkpointing. Parent + PIDray R0/R1 remained close to the Parent\textquotesingle s training duration. These comparisons describe measured cost and do not establish significant timing differences.

\begin{table*}[t]
\footnotesize\centering\caption{Model size, mean training time, and evaluation throughput.}
\begingroup\setlength{\tabcolsep}{2pt}
\begin{tabular*}{\textwidth}{@{\extracolsep{\fill}}l*{3}{c}@{}}
\toprule
\textbf{Experiment}
 & \textbf{Total parameters}
 & \textbf{Training hours, mean}
 & \textbf{End-to-end images/s, mean \ensuremath{\pm} SD}
 \\
\midrule

Mask R-CNN & 63,038,634 & 1.46 & 33.7 \ensuremath{\pm} 4.0 \\
Cascade Mask R-CNN & 96,066,744 & 2.71 & 26.1 \ensuremath{\pm} 1.2 \\
Cascade + AttFPN (Parent) & 97,103,679 & 2.99 & 24.6 \ensuremath{\pm} 0.3 \\
Parent + recall RPN & 97,107,534 & 3.07 & 23.9 \ensuremath{\pm} 3.2 \\
RSPrompter (SAM + LoRA) & 332,667,534 & 4.80 & 14.4 \ensuremath{\pm} 0.3 \\
Occ bilayer SAM\ensuremath{\dagger} & 336,725,874 & 7.21 & 14.3 \ensuremath{\pm} 1.6 \\
Bilayer FCN\ensuremath{\dagger} & 105,173,841 & 4.14 & 21.0 \ensuremath{\pm} 0.3 \\
Parent + PIDray R0/R1 & 98,684,224 & 2.88 & 23.5 \ensuremath{\pm} 0.2 \\
Transformer refiner (ours) & 97,562,062 & 5.22 & 22.3 \ensuremath{\pm} 2.8 \\
MLP refiner (ours) & 97,561,806 & 4.87 & 23.3 \ensuremath{\pm} 2.8 \\
\bottomrule\end{tabular*}\endgroup
\par\smallskip{\raggedright Training hours: three-seed mean. Throughput: mean \ensuremath{\pm} SD across three seeds at evaluation batch size 1, including loading, inference, scoring and writing; not model-only latency. \ensuremath{\dagger} Extra occluder supervision; Occ bilayer SAM used activation checkpointing. Costs are descriptive.\par}
\end{table*}

\section{Discussion}\label{discussion}

On PIXray, the results support H1 for both refiners: integrating point-based refinement increased mean mask AP by 1.29 points for the transformer and 1.97 points for the MLP, with improvements in every evaluated seed under unchanged target-mask supervision. H2 was also supported under the study's \ensuremath{\pm}1-AP equivalence criterion. Although the MLP achieved 0.69 points higher mean AP, the 90\% confidence interval {[}0.49, 0.88{]} remained within those bounds. Thus, attention among selected points provided no observed accuracy advantage in this configuration, although both variants retained contextual features and the Parent's attention.

H3 was supported for the MLP but remained inconclusive for the Transformer; H4 was supported for both on their respective matched-object samples. Joint training and restricted matching limit causal interpretation of the apparent delineation gains. Detection diagnostics established neither improvement nor unchanged performance. Neither refiner improved Dart AP; both increased small-object mask AP but reduced small-object average recall relative to the Parent (Appendix Table A3).

These conclusions apply to complete PIXray configurations; individual contributions of point selection, progressive resolution, shared weights, and additional point loss remain unisolated. Each refiner adds approximately 0.46 million parameters, increases training time, and reduces measured end-to-end throughput; cost comparisons remain descriptive. Generalization is limited by one Parent architecture, three seeds, and earlier design-informing experiments on the same test split. Further limitations include replacing planned Boundary AP with Boundary IoU, approximate p-values, and detection diagnostics added after scoring. Adapted reference implementations, including two with additional occluder supervision, preclude claims of general superiority over the original published systems.

The study focused on PIXray, whose bounding-box and instance-mask annotations support complementary detection and segmentation analyses. Public X-ray segmentation datasets remain limited in diversity and annotation coverage~\cite{ref33}, motivating evaluation across additional benchmarks. Future work should assess generalization on independent datasets and alternative detectors, isolate refinement components through controlled ablations, and compare established refinement methods directly. Tests involving clean baggage, difficult or missed objects, and model-only latency would further clarify practical applicability.

\section{Conclusions}\label{conclusions}

This study evaluated shared point-based mask refinement across cascade stages and resolutions for X-ray baggage instance segmentation. Across three training seeds on the official PIXray split, the transformer and MLP refiners improved mean mask AP over the Parent detector by 1.29 and 1.97 points, respectively, using the original target-mask annotations. The MLP achieved the highest mean mask AP among the ten evaluated experiments, at 72.47\%, while satisfying the study's \ensuremath{\pm}1-AP equivalence criterion relative to the transformer. These findings support point-based refinement within the tested architecture and show that additional self-attention among selected points was not required to obtain the observed gains.

The mask-quality analyses point mainly to improved delineation: both refiners increased boundary IoU on commonly matched objects, while only the MLP supported the AP\textsubscript{75}-minus-AP\textsubscript{50} gain contrast. Detection diagnostics showed no corresponding positive gain. The improvements came with a small increase in parameter count, longer training, and lower measured end-to-end throughput. The results therefore show an accuracy--cost trade-off for the evaluated PIXray configurations. Evaluation on independent datasets is needed to assess generalization. Direct comparisons with established refiners and controlled component ablations should clarify relative performance and the contributions of individual components. Clean-image evaluation and model-only latency measurements remain necessary to assess operational suitability.

\appendix\label{appendix}
\setcounter{table}{0}\renewcommand{\thetable}{A\arabic{table}}\renewcommand{\theHtable}{appendix.\arabic{table}}

\subsection{Supporting numerical results}\label{a.1-supporting-numerical-results}

\begin{table*}[t]
\footnotesize\centering\caption{Paired per-class mask-AP differences relative to the Parent.}
\begingroup\setlength{\tabcolsep}{2pt}
\begin{tabular*}{\textwidth}{@{\extracolsep{\fill}}l*{6}{c}@{}}
\toprule
\textbf{Class}
 & \textbf{T \ensuremath{\Delta} {[}95\% CI{]}}
 & \textbf{p}
 & \textbf{Holm p}
 & \textbf{M \ensuremath{\Delta} {[}95\% CI{]}}
 & \textbf{p}
 & \textbf{Holm p}
 \\
\midrule

Dart & \ensuremath{-}0.3 {[}\ensuremath{-}1.6, +0.7{]} & 0.576 & 1 & \ensuremath{-}0.3 {[}\ensuremath{-}1.0, +0.6{]} & 0.49 & 1 \\
Saw blade & \ensuremath{-}0.4 {[}\ensuremath{-}2.3, +1.3{]} & 0.683 & 1 & +4.3 {[}+2.7, +5.7{]} & \textless0.001 & \textless0.001 \\
Knife & +0.5 {[}\ensuremath{-}0.6, +1.6{]} & 0.414 & 1 & +1.3 {[}+0.0, +2.5{]} & 0.0309 & 0.216 \\
Screwdriver & +3.3 {[}+1.8, +4.6{]} & \textless0.001 & \textless0.001 & +2.7 {[}+1.4, +3.9{]} & \textless0.001 & \textless0.001 \\
Pliers & +0.7 {[}\ensuremath{-}1.2, +2.3{]} & 0.43 & 1 & +4.2 {[}+3.2, +5.5{]} & \textless0.001 & \textless0.001 \\
Lighter & +1.0 {[}+0.2, +1.9{]} & 0.0222 & 0.155 & +0.5 {[}\ensuremath{-}0.4, +1.4{]} & 0.321 & 1 \\
Razor blade & \ensuremath{-}0.1 {[}\ensuremath{-}1.0, +0.8{]} & 0.858 & 1 & +0.3 {[}\ensuremath{-}0.9, +1.5{]} & 0.669 & 1 \\
Wrench & +1.9 {[}+0.8, +2.9{]} & \textless0.001 & 0.00542 & +2.8 {[}+1.9, +3.7{]} & \textless0.001 & \textless0.001 \\
Scissors & +1.5 {[}+0.7, +2.4{]} & \textless0.001 & 0.00643 & +1.6 {[}+0.9, +2.4{]} & \textless0.001 & \textless0.001 \\
Fireworks & +1.3 {[}+0.3, +2.4{]} & 0.0167 & 0.134 & +1.1 {[}+0.0, +2.3{]} & 0.0614 & 0.307 \\
Battery & \ensuremath{\approx}0.0 {[}\ensuremath{-}0.7, +0.8{]} & 0.999 & 1 & +0.1 {[}\ensuremath{-}0.7, +1.0{]} & 0.848 & 1 \\
Gun & +1.2 {[}+0.3, +1.9{]} & 0.00329 & 0.0296 & +1.3 {[}+0.5, +2.1{]} & 0.00168 & 0.0135 \\
Hammer & +1.0 {[}+0.4, +1.6{]} & 0.00101 & 0.0101 & +0.9 {[}\ensuremath{-}0.1, +1.5{]} & 0.0385 & 0.231 \\
Pressure vessel & +2.7 {[}+1.6, +3.8{]} & \textless0.001 & \textless0.001 & +3.3 {[}+2.3, +4.4{]} & \textless0.001 & \textless0.001 \\
Bat & +5.0 {[}+4.2, +6.0{]} & \textless0.001 & \textless0.001 & +5.4 {[}+4.7, +6.2{]} & \textless0.001 & \textless0.001 \\
\bottomrule\end{tabular*}\endgroup
\par\smallskip{\raggedright AP points; T = transformer \ensuremath{-} Parent, M = MLP \ensuremath{-} Parent. p: CI-normal approximation; Holm across 15 classes per contrast. CIs are marginal. Transformer Battery is slightly negative before rounding. Boldface in Table 5 does not indicate statistical significance.\par}
\end{table*}

\begin{table*}[t]\centering
\begin{minipage}[t]{.49\textwidth}

\footnotesize\centering\caption{Mask AP for each training seed.}
\begingroup\setlength{\tabcolsep}{2pt}
\begin{tabular*}{\linewidth}{@{\extracolsep{\fill}}l*{3}{c}@{}}
\toprule
\textbf{Experiment}
 & \textbf{Seed 42}
 & \textbf{Seed 123}
 & \textbf{Seed 2026}
 \\
\midrule

Mask R-CNN & 68.1049 & 68.1415 & 68.2057 \\
Cascade Mask R-CNN & 70.1905 & 70.2353 & 70.1653 \\
Cascade + AttFPN (Parent) & 70.4832 & 70.4431 & 70.5644 \\
Parent + recall RPN & 70.4843 & 70.1862 & 70.0532 \\
RSPrompter (SAM + LoRA) & 62.5149 & 62.3249 & 62.5245 \\
Occ bilayer SAM\ensuremath{\dagger} & 62.7329 & 62.6993 & 62.5528 \\
Bilayer FCN\ensuremath{\dagger} & 70.7241 & 70.6629 & 70.5585 \\
Parent + PIDray R0/R1 & 70.4018 & 70.4481 & 70.2191 \\
Transformer refiner (ours) & 71.8995 & 71.5642 & 71.8927 \\
MLP refiner (ours) & 72.5409 & 72.3802 & 72.4922 \\
\bottomrule\end{tabular*}\endgroup
\par\smallskip{\raggedright AP on the 0--100 scale, rounded to four decimals; Table 2 uses unrounded data. \ensuremath{\dagger} Extra occluder supervision.\par}
\end{minipage}\hfill\begin{minipage}[t]{.49\textwidth}

\footnotesize\centering\caption{Mask AP and average recall by object size.}
\begingroup\setlength{\tabcolsep}{2pt}
\begin{tabular*}{\linewidth}{@{\extracolsep{\fill}}l*{6}{c}@{}}
\toprule
\textbf{Experiment}
 & \textbf{AP\textsubscript{S}}
 & \textbf{AP\textsubscript{M}}
 & \textbf{AP\textsubscript{L}}
 & \textbf{AR\textsubscript{S}}
 & \textbf{AR\textsubscript{M}}
 & \textbf{AR\textsubscript{L}}
 \\
\midrule

Mask R-CNN & 44.2 & 54.9 & 68.6 & 57.3 & 66.6 & 69.4 \\
Cascade Mask R-CNN & 42.7 & 58.2 & 70.1 & 58.0 & 67.7 & 70.8 \\
Cascade + AttFPN (Parent) & 43.8 & 59.1 & 69.5 & 60.9 & 67.9 & 70.1 \\
Parent + recall RPN & 45.5 & 58.7 & 69.8 & 62.0 & 68.0 & 70.4 \\
RSPrompter (SAM + LoRA) & 32.3 & 47.7 & 67.1 & 43.6 & 57.6 & 67.7 \\
Occ bilayer SAM\ensuremath{\dagger} & 33.7 & 48.0 & 64.5 & 46.3 & 57.5 & 65.1 \\
Bilayer FCN\ensuremath{\dagger} & 43.7 & 59.0 & 70.0 & 62.3 & 68.3 & 70.7 \\
Parent + PIDray R0/R1 & 41.1 & 58.8 & 69.9 & 58.6 & 67.8 & 70.6 \\
Transformer refiner (ours) & 46.5 & 59.8 & 72.0 & 57.6 & 68.5 & 72.6 \\
MLP refiner (ours) & 46.2 & 60.5 & 72.6 & 58.8 & 68.6 & 73.2 \\
\bottomrule\end{tabular*}\endgroup
\par\smallskip{\raggedright Three-seed means (\%). S, M and L use COCO size ranges with cut points at 32\textsuperscript{2} and 96\textsuperscript{2} pixels\textsuperscript{2}; ground-truth size follows PIXray's stored box-area field. AP and AR average over mask-IoU thresholds 0.50--0.95; size-specific AR uses maxDets = 100. \ensuremath{\dagger} Additional occluder supervision.\par}
\end{minipage}\end{table*}

Figure 8 displays seed-42 results selected by recorded performance-based rules from objects with boxes \ensuremath{\geq} 24 pixels per side in images containing \ensuremath{\leq} 6 annotated objects. Except for the missed-object example, candidates were matched by all five displayed experiments in every seed. Improvement cases favored both refiners over the best displayed reference in all seeds (refiner IoU \ensuremath{\geq} 0.70), including a difficult-class example. The remaining cases show a consistent Wrench regression, a pressure vessel missed by all five experiments, and a Dart with the lowest best-of-five seed-42 IoU among eligible Darts. These selected cases illustrate behavior rather than estimate its prevalence.

\subsection{Additional descriptive metrics}\label{a.2-additional-descriptive-metrics}

Tables A3, A4, and A5 summarize additional outcomes for all ten experiments. Values are arithmetic means over seeds 42, 123, and 2026 on the same 1,486-image test set and are reported on a 0--100 scale. These comparisons are descriptive; no additional hypothesis tests or confidence intervals are introduced.

Both refiners had higher AP than the Parent in each size group. The MLP had the highest medium- and large-object AP, while the transformer had the highest small-object AP. Small-object AR was lower for both refiners than for the Parent, so the AP improvement did not extend to every recall measure.

\begin{table*}[t]\centering
\begin{minipage}[t]{.49\textwidth}
\footnotesize\centering\caption{Mask precision, recall and F1 at a fixed score threshold.}
\begingroup\setlength{\tabcolsep}{2pt}
\begin{tabular*}{\linewidth}{@{\extracolsep{\fill}}l*{6}{c}@{}}
\toprule
\textbf{Experiment}
 & \textbf{P\textsubscript{50}}
 & \textbf{R\textsubscript{50}}
 & \textbf{F1\textsubscript{50}}
 & \textbf{P\textsubscript{75}}
 & \textbf{R\textsubscript{75}}
 & \textbf{F1\textsubscript{75}}
 \\
\midrule

Mask R-CNN & 94.7 & 93.3 & 94.0 & 82.9 & 81.6 & 82.3 \\
Cascade Mask R-CNN & 96.9 & 93.0 & 94.9 & 86.6 & 83.0 & 84.8 \\
Cascade + AttFPN (Parent) & 97.2 & 93.3 & 95.2 & 86.9 & 83.4 & 85.1 \\
Parent + recall RPN & 96.8 & 93.6 & 95.2 & 86.0 & 83.1 & 84.5 \\
RSPrompter (SAM + LoRA) & 89.4 & 92.9 & 91.1 & 71.2 & 73.9 & 72.5 \\
Occ bilayer SAM\ensuremath{\dagger} & 89.4 & 92.9 & 91.1 & 71.1 & 73.9 & 72.5 \\
Bilayer FCN\ensuremath{\dagger} & 96.7 & 93.7 & 95.2 & 86.4 & 83.7 & 85.1 \\
Parent + PIDray R0/R1 & 97.2 & 93.3 & 95.2 & 86.7 & 83.2 & 84.9 \\
Transformer refiner (ours) & 97.3 & 93.1 & 95.1 & 86.9 & 83.2 & 85.0 \\
MLP refiner (ours) & 97.5 & 93.3 & 95.4 & 88.0 & 84.3 & 86.1 \\
\bottomrule\end{tabular*}\endgroup
\par\smallskip{\raggedright Three-seed means (\%) at score \ensuremath{\geq} 0.50; subscripts denote mask-IoU thresholds. Matching uses identifier-zero-safe COCO greedy, score-ordered, class-wise assignment; P, R and F1 are micro-averaged within each seed, then averaged across seeds, with recall over all 4,508 objects. \ensuremath{\dagger} Extra occluder supervision.\par}

\end{minipage}\hfill\begin{minipage}[t]{.49\textwidth}
\footnotesize\centering\caption{Object-level mask overlap and detection coverage.}
\begingroup\setlength{\tabcolsep}{2pt}
\begin{tabularx}{\linewidth}{>{\raggedright\arraybackslash}p{.354\linewidth}>{\centering\arraybackslash}p{.14\linewidth}>{\centering\arraybackslash}p{.136\linewidth}>{\centering\arraybackslash}p{.12\linewidth}>{\centering\arraybackslash}p{.17\linewidth}}
\toprule
\textbf{Experiment}
 & \textbf{Coverage}
 & \parbox[t]{\linewidth}{\centering\bfseries Mean\\mask IoU}
 & \textbf{Mean Dice}
 & \parbox[t]{\linewidth}{\centering\bfseries Matched\\masks with\\IoU\,\ensuremath{\geq}\,0.90\\(\%)}
 \\
\midrule

Mask R-CNN & 94.2 & 85.3 & 91.7 & 42.5 \\
Cascade Mask R-CNN & 93.7 & 86.0 & 92.2 & 45.5 \\
Cascade + AttFPN (Parent) & 94.0 & 86.1 & 92.2 & 45.6 \\
Parent + recall RPN & 94.2 & 86.0 & 92.1 & 45.1 \\
RSPrompter (SAM + LoRA) & 95.1 & 81.5 & 89.2 & 26.4 \\
Occ bilayer SAM\ensuremath{\dagger} & 94.9 & 81.6 & 89.3 & 26.8 \\
Bilayer FCN\ensuremath{\dagger} & 94.4 & 86.1 & 92.2 & 45.5 \\
Parent + PIDray R0/R1 & 93.9 & 86.1 & 92.2 & 45.4 \\
Transformer refiner (ours) & 93.7 & 86.7 & 92.6 & 49.8 \\
MLP refiner (ours) & 93.8 & 87.2 & 92.9 & 50.3 \\
\bottomrule\end{tabularx}\endgroup
\par\smallskip{\raggedright Three-seed means (\%), with one-to-one same-class box-IoU matching at score \ensuremath{\geq} 0.50 and box IoU \ensuremath{\geq} 0.50; coverage uses all 4,508 objects. Overlap scores use each experiment's own matched objects, unlike Table A4's mask-IoU matching and Table 4's paired common-match samples; \ensuremath{\dagger} extra occluder supervision. Coverage is not interchangeable with COCO AR\textsubscript{100} or AP\textsubscript{50}.\par}
\end{minipage}
\end{table*}

At mask IoU 0.75, F1 was 86.1\% for the MLP, 85.0\% for the transformer, and 85.1\% for the Parent. At IoU 0.50, the corresponding values were 95.4\%, 95.1\%, and 95.2\%. The descriptive advantage at this operating point was therefore clearer for the MLP; the transformer's AP gain over the Parent did not imply a higher F1 at either displayed threshold.

The refiners had higher descriptive mean overlap scores than the Parent. The proportion of matched masks reaching IoU 0.90 rose from 45.6\% for the Parent to 49.8\% for the transformer and 50.3\% for the MLP, while coverage remained slightly lower. Because matched sets differ between experiments, these absolute summaries complement rather than replace the paired comparisons.

\clearpage\twocolumn


\begin{thebibliography}{00}
\bibitem{ref1}
Akcay, S. \& Breckon, T. P. Towards automatic threat detection: A survey of advances of deep learning within X-ray security imaging. \emph{Pattern Recognit.} 122, 108245 (2022).
\bibitem{ref2}
Zhang, L., Jiang, L., Ji, R. \& Fan, H. PIDray: A large-scale X-ray benchmark for real-world prohibited item detection. \emph{Int. J. Comput. Vis.} 131, 3170--3192 (2023).
\bibitem{ref3}
Ma, B., Jia, T., Su, M., Jia, X., Chen, D. \& Zhang, Y. Automated segmentation of prohibited items in X-ray baggage images using dense de-overlap attention snake. \emph{IEEE Trans. Multimedia} 25, 4374--4386 (2023).
\bibitem{ref4}
He, K., Gkioxari, G., Dollár, P. \& Girshick, R. Mask R-CNN. \emph{Proc. IEEE Int. Conf. Comput. Vis.} (2017).
\bibitem{ref5}
Cai, Z. \& Vasconcelos, N. Cascade R-CNN: Delving into high quality object detection. \emph{Proc. IEEE Conf. Comput. Vis. Pattern Recognit.} (2018).
\bibitem{ref6}
Ren, Y., Li, R., Liu, L. \& Chen, C. Prohibited items segmentation via occlusion-aware bilayer modeling. \emph{Proc. IEEE Int. Conf. Multimedia Expo} (2025).
\bibitem{ref7}
Kirillov, A., Wu, Y., He, K. \& Girshick, R. PointRend: Image segmentation as rendering. \emph{Proc. IEEE/CVF Conf. Comput. Vis. Pattern Recognit.} (2020).
\bibitem{ref8}
Zhang, G. et al. RefineMask: Towards high-quality instance segmentation with fine-grained features. \emph{Proc. IEEE/CVF Conf. Comput. Vis. Pattern Recognit.} (2021).
\bibitem{ref9}
Ke, L. et al. Mask Transfiner for high-quality instance segmentation. \emph{Proc. IEEE/CVF Conf. Comput. Vis. Pattern Recognit.} (2022).
\bibitem{ref10}
Tao, R. et al. Dual-view X-ray detection: Can AI detect prohibited items from dual-view X-ray images like humans? \emph{Proc. IEEE/CVF Conf. Comput. Vis. Pattern Recognit.} 10338--10347 (2025).
\bibitem{ref11}
Zhang, S., Zhu, D., Yung, K. \& Ip, A. W. H. Balanced X-ray security dataset and enhanced YOLO for contraband detection. \emph{Sci. Data} 12, 2041 (2025).
\bibitem{ref12}
Zhang, H., Zhao, Z. \& Yang, J. Attention-based prohibited item detection in X-ray images during security checking. \emph{IET Image Process.} 18, 1119--1131 (2024).
\bibitem{ref13}
Nasim, A. et al. Regressor driven incremental instance segmentation for contraband items identification. \emph{Eng. Appl. Artif. Intell.} 157, 111421 (2025).
\bibitem{ref14}
Li, M. et al. AO-DETR: Anti-overlapping DETR for X-ray prohibited items detection. \emph{arXiv} 2403.04309 (2024).
\bibitem{ref15}
Li, M. et al. MMCL: Correcting content query distributions for improved anti-overlapping X-ray object detection. \emph{arXiv} 2406.03176v2 (2025).
\bibitem{ref16}
Wang, Z.-W., Shen, Y., Li, W. \& Yuan, Z. Prohibited item detection in X-ray images based on refined surface perception. \emph{J. Vis. Commun. Image Represent.} 119, 104864 (2026).
\bibitem{ref17}
Lin, S., Jia, T., Wang, H., Ma, B., Li, M. \& Chen, D. Open-vocabulary X-ray prohibited item detection via fine-tuning CLIP. \emph{arXiv} 2406.10961 (2024).
\bibitem{ref18}
Chen, K. et al. RSPrompter: Learning to prompt for remote sensing instance segmentation based on visual foundation model. \emph{IEEE Trans. Geosci. Remote Sens.} 62, 1--17 (2024).
\bibitem{ref19}
Ravi, N. et al. SAM 2: Segment anything in images and videos. \emph{arXiv} 2408.00714 (2024).
\bibitem{ref20}
Lin, Y. et al. SAMRefiner: Taming segment anything model for universal mask refinement. \emph{Proc. Int. Conf. Learn. Represent.} (2025).
\bibitem{ref21}
He, K., Zhang, X., Ren, S. \& Sun, J. Deep residual learning for image recognition. \emph{Proc. IEEE Conf. Comput. Vis. Pattern Recognit.} (2016).
\bibitem{ref22}
Lin, T.-Y. et al. Feature pyramid networks for object detection. \emph{Proc. IEEE Conf. Comput. Vis. Pattern Recognit.} (2017).
\bibitem{ref23}
Kirillov, A. et al. Segment anything. \emph{Proc. IEEE/CVF Int. Conf. Comput. Vis.} (2023).
\bibitem{ref24}
Hu, E. J. et al. LoRA: Low-rank adaptation of large language models. \emph{Proc. Int. Conf. Learn. Represent.} (2022).
\bibitem{ref25}
Ke, L., Tai, Y.-W. \& Tang, C.-K. Deep occlusion-aware instance segmentation with overlapping bilayers. \emph{Proc. IEEE/CVF Conf. Comput. Vis. Pattern Recognit.} (2021).
\bibitem{ref26}
Chen, K. et al. MMDetection: Open MMLab detection toolbox and benchmark. \emph{arXiv} 1906.07155 (2019).
\bibitem{ref27}
Lin, T.-Y. et al. Microsoft COCO: Common objects in context. \emph{Proc. Eur. Conf. Comput. Vis.} (2014).
\bibitem{ref28}
Efron, B. Bootstrap methods: Another look at the jackknife. \emph{Ann. Stat.} 7, 1--26 (1979).
\bibitem{ref29}
Phipson, B. \& Smyth, G. K. Permutation p-values should never be zero: Calculating exact p-values when permutations are randomly drawn. \emph{Stat. Appl. Genet. Mol. Biol.} 9, 39 (2010).
\bibitem{ref30}
Schuirmann, D. J. A comparison of the two one-sided tests procedure and the power approach for assessing the equivalence of average bioavailability. \emph{J. Pharmacokinet. Biopharm.} 15, 657--680 (1987).
\bibitem{ref31}
Cheng, B. et al. Boundary IoU: Improving object-centric image segmentation evaluation. \emph{Proc. IEEE/CVF Conf. Comput. Vis. Pattern Recognit.} (2021).
\bibitem{ref32}
Holm, S. A simple sequentially rejective multiple test procedure. \emph{Scand. J. Stat.} 6, 65--70 (1979).
\bibitem{ref33}
Gao et al. XSeg: A large-scale X-ray contraband segmentation benchmark for real-world security screening. Proc. IEEE/CVF Conf. Comput. Vis. Pattern Recognit. (2026).
\end{thebibliography}
\end{document}